\documentclass[10pt,twocolumn,letterpaper]{article}

\usepackage{iccv}              

\definecolor{iccvblue}{rgb}{0.21,0.49,0.74}
\usepackage[pagebackref,breaklinks,colorlinks,allcolors=iccvblue]{hyperref}
\usepackage{multirow}
\usepackage{tabularx}
\usepackage{amssymb}  
\usepackage{pifont}   
\usepackage{float}

\newcommand{\xmark}{\ding{55}} 

\def\paperID{CVAM-13} 
\def\confName{ICCV}
\def\confYear{2025}

\title{Seeing the Signs: A Survey of Edge-Deployable OCR Models for Billboard Visibility Analysis}

\author{
Maciej Szankin \quad Vidhyananth Venkatasamy \quad Lihang Ying\\
SiMa.ai\\
333 W San Carlos St \#1100, San Jose, CA 95110\\
{\tt\small \{maciej.szankin, vidhyananth.venkatasamy, lihang.ying\}@sima.ai}
}

\begin{document}
\maketitle

\begin{abstract}
Outdoor advertisements remain a critical medium for modern marketing, yet accurately verifying billboard text visibility under real-world conditions is still challenging. Traditional Optical Character Recognition (OCR) pipelines excel at cropped text recognition but often struggle with complex outdoor scenes, varying fonts, and weather-induced visual noise. Recently, multimodal Vision-Language Models (VLMs) have emerged as promising alternatives, offering end-to-end scene understanding with no explicit detection step. This work systematically benchmarks representative VLMs—including Qwen 2.5 VL 3B, InternVL3, and SmolVLM2—against a compact CNN-based OCR baseline (PaddleOCRv4) across two public datasets (ICDAR 2015 and SVT), augmented with synthetic weather distortions to simulate realistic degradation. Our results reveal that while selected VLMs excel at holistic scene reasoning, lightweight CNN pipelines still achieve competitive accuracy for cropped text at a fraction of the computational cost—an important consideration for edge deployment. To foster future research, we release our weather-augmented benchmark and evaluation code publicly $<$ link provided upon acceptance $>$.
\end{abstract}

\section{Introduction}

In modern marketing, outdoor advertisements such as billboards, banners, and shop signs remain essential for brand visibility and regional targeting. As campaigns scale across geographies and vendors, ensuring that physical ads are correctly displayed, legible, and brand-compliant becomes a critical challenge. Optical Character Recognition (OCR) offers a scalable, automated means to extract textual content from physical ad media for validation and analysis.

OCR helps detect whether a billboard contains the correct product name, promotional language, or legal disclaimers, preventing deployment errors. In digital-out-of-home (DOOH) advertising, OCR verifies dynamic ad playback content against scheduling data. By analyzing OCR-extracted text, advertisers can also assess whether messaging was clearly visible and link this to downstream digital metrics such as click-throughs or search volume.

OCR also enables personalized advertising systems - for example, using detected text cues like license plates or signage to trigger tailored content. Retail environments benefit too, using OCR to recognize product labels, shelf tags, or brand logos for tasks ranging from promotion auditing to inventory tracking. Across these use cases, OCR facilitates both real-time ad intelligence and long-term campaign analytics.

Emerging benchmarks and challenges in computer vision now target urban signage recognition, underscoring the field’s growing focus on automating ad tracking at scale. However, these real-world scenes often involve stylized fonts, curved layouts, occlusions, or variable lighting - all of which degrade the accuracy of traditional OCR. As a result, advanced multimodal methods are increasingly explored for their ability to reason about text in context-rich visual environments.

Vision-Language Models (VLMs), pretrained on paired image-text datasets, offer improved robustness in such scenarios. Instead of relying solely on explicit text detection and recognition, these models can infer the presence and semantics of text as part of a broader image understanding task. Their application to advertising OCR remains an emerging but promising frontier.

When designing OCR systems, it is essential to consider the diverse range of use cases and visual variations in text. OCR for scanned documents requires high accuracy with serif fonts and structured layouts \cite{smith2007overview}, whereas OCR for street signs or product labels must handle complex backgrounds, varying lighting, and distorted angles \cite{jaderberg2016reading}. Real-time license plate recognition demands speed and robustness to motion blur \cite{silva2018license}, while historical document digitization emphasizes tolerance to degradation and non-standard scripts \cite{fischer2012lexicon}. In advertising, OCR must cope with stylized fonts, creative layouts, varying text orientations, and overlays on visually rich or cluttered backgrounds \cite{liu2020abcnet}. Variation between handwritten and printed text adds further complexity, requiring specialized models and often sequence-based approaches for effective recognition \cite{bluche2017scan}. Effective OCR systems must therefore be adaptable, often incorporating preprocessing steps like image enhancement, text localization, and geometric normalization to handle diverse visual challenges.

This study investigates the efficacy of different OCR techniques for analyzing billboard text visibility in marketing applications, with emphasis on automation, scalability, and edge deployment. The goal is to evaluate how well machine learning models can extract meaningful text from real-world advertising scenes - both synthetic and urban - under conditions like occlusion, distortion, and stylized fonts.

We compare traditional OCR pipelines - such as Convolutional Neural Network-based (CNN) systems like PaddleOCR \cite{du2020pp} - with modern VLMs, including Gemma 3 4B \cite{team2025gemma} and Qwen2.5-VL 3B \cite{Qwen2VL}. These models are evaluated on both cropped text regions and holistic full-image contexts to benchmark their real-world usability across multiple advertising scenarios.

To support robustness research, we extend standard benchmark datasets - ICDAR 2015 (IC15) \cite{karatzas2015icdar} and Street View Text (SVT) \cite{yin2015big}) - with realistic weather augmentations, including synthetic rain, fog, and combined conditions at multiple severity levels. This expanded dataset helps assess model behavior under challenging real-world degradations and is available publicly to encourage reproducibility and further work on OCR resilience.

A key focus of our analysis is deployment feasibility for edge-based retail applications, including the trade-offs between model accuracy, inference latency, and memory consumption. Through extensive experiments, we identify behaviors such as hallucination in text generation and robustness to visual variation, and we discuss how these affect billboard interpretation quality.

Ultimately, this work contributes to the growing intersection of computer vision, urban computing, and marketing technology, showing how machine perception can drive smarter, more responsive advertising systems.

\section{Related Work}

Recent advances in OCR are driven by three major families of models: CNN-based, Transformer-based, and large VLMs. Traditional CNN-based approaches such as CRNN \cite{shi2016crnn}, Rosetta \cite{borisyuk2018rosetta}, and TextFuseNet \cite{ye2020textfusenet} extract spatial features using convolutions and model sequential dependencies with Recurrent Neural Networks (RNNs) or attention mechanisms, decoding via Connectionist Temporal Classification (CTC) or attention. PaddleOCR, a popular open-source OCR system, evolved through v4 and v5 \cite{wang2022ppocrv4, paddleocr2024v5}, integrating lightweight CNN backbones like LCNet and modern decoders like PARSeq. SVTRv2 \cite{du2024svtrv2} blends CNN and Transformer features in a streamlined architecture with CTC loss, outperforming encoder-decoder alternatives in both speed and accuracy.

Transformer-based models offer global context and flexible decoding. PARSeq \cite{bautista2022parseq} introduces a unified framework capable of both autoregressive and non-autoregressive sequence prediction. DAN \cite{wang2020dan} improves robustness by decoupling attention alignment from decoding history, enhancing performance on irregular text. These models, while heavier than CNN+CTC pipelines, yield superior accuracy on complex scripts and degraded images.

Hybrid systems like OpenOCR \cite{openocr2025}, which integrates SVTRv2 with modular decoding, combine the efficiency of CNNs with the modeling power of Transformers. OpenOCR achieves a 4.5\% improvement over PaddleOCR on multilingual benchmarks and serves as a flexible open-source pipeline for detection, recognition, and layout parsing in practical deployments.

Recent studies extend these themes into real-world advertising and retail applications. For example, shop sign recognition competitions highlight OCR’s role in analyzing commercial visibility in street scenes \cite{tang2023character}. Some frameworks adopt meta-learning for generalist multimodal agents that adapt to OCR tasks with limited examples, bridging cognitive science and AI for real-world applications.

Smart billboard systems have shown how OCR enables personalized ad targeting in physical retail \cite{sumer22new}. For instance, license plate recognition can identify vehicle owners and deliver tailored ads by linking recognized text to customer databases. Such pipelines demonstrate practical OCR use in enhancing consumer engagement.

Scene text understanding has also been applied to link product information with campaign monitoring. Unitail \cite{chen2022unitail} is a large-scale retail benchmark combining product detection with OCR for 1.8 million product instances, 30K text regions, and 21K transcriptions. Its experiments show that OCR-enhanced matching significantly improves accuracy when linking detected products to galleries, and its dataset and evaluation server are publicly available.

OCR has supported visual question answering (VQA) frameworks that interpret advertisements for semantic ad analysis. AdQuestA \cite{choudhary2025adquesta} combines textual and visual ad context, introducing ADVQA, the first benchmark for ad-related VQA. Using Retrieval Augmented Generation (RAG) with OCR-extracted text, AdQuestA outperforms large language models like GPT-4 in ad understanding, demonstrating OCR’s role in multimodal reasoning.

Finally, multimodal models assisted by OCR have been used to detect unauthorized or illegal ads by analyzing both image and text cues \cite{zhang2024identification}. A dual-stage framework combines fine-tuned CLIP and PaddleOCRv4 to classify and verify ads against banned keyword databases, achieving 93.5\% accuracy and demonstrating the critical role of OCR in visual regulatory compliance.

\section{Methodology}

\subsection{Dataset Augmentation for Weather Robustness}

To evaluate the performance and robustness of the selected OCR models, we utilize two publicly available scene text datasets: ICDAR 2015 (IC15) and Street View Text (SVT). These datasets serve as the basis for our evaluation pipeline and are representative of real-world outdoor text in complex urban scenes. Both include word-level annotations and offer significant variability in terms of font, background, resolution, and layout. To simulate challenging environmental conditions, we further augment these datasets with synthetic weather transformations, as described in Section \ref{subsec:augmentation}

\begin{figure}[h]
\centering
\begin{subfigure}{0.48\linewidth}
  \centering
  \includegraphics[width=\linewidth]{imgs/icdar_img_498_original.jpg}
  \caption{ICDAR 2015}
  \label{fig:sample-icdar}
\end{subfigure}
\hfill
\begin{subfigure}{0.48\linewidth}
  \centering
  \includegraphics[width=\linewidth]{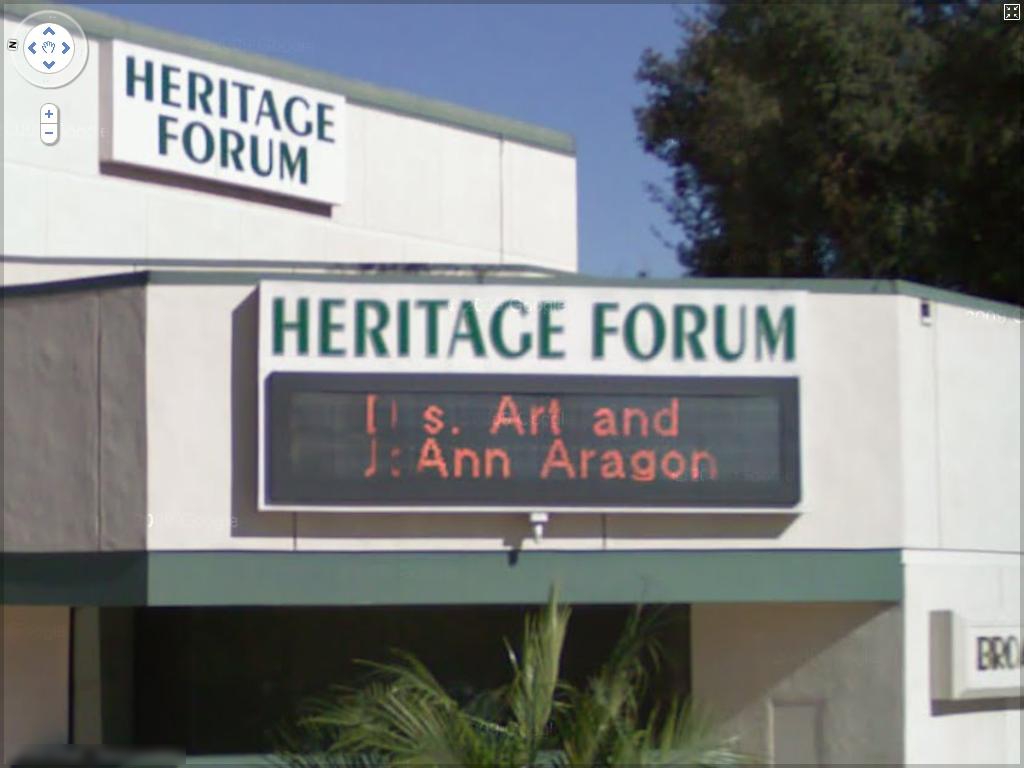}
  \caption{SVT}
  \label{fig:sample-svt}
\end{subfigure}
\caption{Example images from the two datasets used in this study. Left: multilingual street sign from ICDAR 2015. Right: storefront text from SVT.}
\label{fig:sample-datasets}
\end{figure}

\subsubsection{ICDAR 2015 (IC15)}
The ICDAR 2015 Text Reading in the Wild dataset was introduced as part of a multilingual OCR competition and is designed to benchmark text detection and recognition algorithms on both English and Chinese scripts. It consists of natural scene images with text captured in diverse conditions, exhibiting high variability in orientation, font, and background clutter. The dataset emphasizes the need for multilingual support and real-world complexity, reflecting practical OCR challenges beyond Latin-script centric benchmarks. In our work, we focus on English-language recognition and skip annotations not based on the Latin alphabet. In our work, we use the test subset of the ICDAR 2015 dataset, which contains 500 images. An example image from the dataset is presented on \ref{fig:sample-icdar}.

\subsubsection{Street View Text}
The SVT dataset was collected from Google Street View imagery and primarily contains English text extracted from storefronts and signage in outdoor urban environments. It is characterized by significant variation in resolution, text quality, and scene composition. SVT includes only word-level annotations and is commonly used for lexicon-driven word recognition and word spotting tasks. Its strong ties to real-world business signage make it particularly relevant for evaluating OCR systems in advertising-related applications. In our work, we use the test subset of the SVT dataset, which contains 249 images. An example image from the dataset is presented on \ref{fig:sample-svt}.

\subsubsection{Synthetic Weather Augmentation}
\label{subsec:augmentation}

To simulate real-world degradation conditions and evaluate model robustness under adverse environments, we apply a series of synthetic weather augmentations to the base datasets. Specifically, we introduce three weather scenarios: rain, fog, and a combined rain-and-fog condition. Each scenario is further divided into three severity levels: light, medium, and heavy.

\paragraph{Rain.}
The rain effect is applied by procedurally generating slanted raindrop streaks of varying opacity and blending them into the original image. Drops are rendered using randomized coordinates and line drawing with a slight motion blur to simulate rainfall streaks. A subtle blue tint is added to emulate atmospheric light scattering typical in rainy weather. The severity of rain is controlled by the number of generated raindrops:

\begin{itemize}
    \item Light - 800 drops
    \item Medium: 1600 drops
    \item Heavy: 2400 drops
\end{itemize}

\paragraph{Fog.}
The fog effect is generated by compositing multi-octave noise layers to produce a cloud-like texture. This texture is blurred at multiple scales and blended with the original image using a gradient mask, which adds more fog density toward the top of the image to simulate natural fog behavior. RGB channel perturbations are used to introduce realistic bluish-gray coloration. The severity of fog is controlled by the blending coefficient: 

\begin{itemize}
    \item Light: $\alpha = 0.5$
    \item Medium: $\alpha = 0.7$
    \item Heavy: $\alpha = 0.9$.
\end{itemize}

\paragraph{Combined Rain and Fog.}
For the third condition, both augmentations are applied sequentially—rain is first added to the image, followed by fog. This simulates more extreme visibility degradation and tests compound robustness.

These augmentations are applied uniformly across both datasets (ICDAR 2015 and SVT), and each image is transformed independently across all three weather conditions and three intensity levels. This results in nine augmented variants per original image, substantially increasing evaluation diversity while maintaining the original annotation structure.

\begin{figure*}[h]
\centering
\begin{subfigure}{0.48\linewidth}
  \centering
  \includegraphics[width=\linewidth]{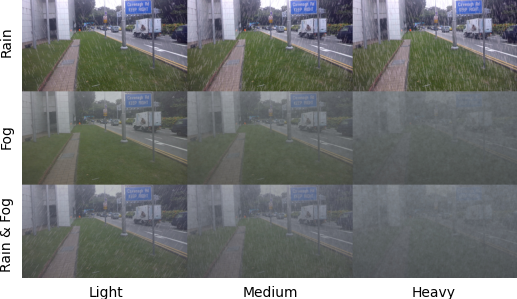}
  \caption{ICDAR 2015}
  \label{fig:augmented-icdar}
\end{subfigure}
\hfill
\begin{subfigure}{0.48\linewidth}
  \centering
  \includegraphics[width=\linewidth]{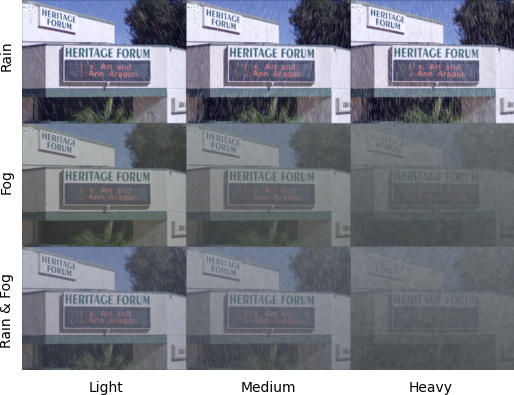}
  \caption{SVT}
  \label{fig:augmented-svt}
\end{subfigure}
\caption{Visual effects of weather-based augmentations applied to an image from ICDAR 2015 and SVT datasets. Each row shows a different weather condition: \textit{Rain}, \textit{Fog} and a combination of \textit{Rain \& Fog}. Columns represent increasing augmentation intensity: \textit{Light}, \textit{Medium} and \textit{Heavy}. These augmentations simulate challenging conditions for OCR robustness evaluation.}
\label{fig:augmented-datasets}
\end{figure*}

Resulting augmented datasets contain 2490 and 5000 images for SVT and ICDAR 2015, respectively. The effects of these augmentations on sample images from the original datasets are illustrated in Fig. \ref{fig:augmented-datasets}.



\subsection{Model Selection}
In this work, we evaluate a representative set of models spanning two categories relevant to scene text recognition in urban environments. The first category includes traditional CNN-based OCR systems, represented by PaddleOCRv4. We focus specifically on its recognition module, which employs a lightweight CRNN architecture well-suited for edge deployment. Detection is excluded to enable a clearer comparison with end-to-end models.

The second category comprises Vision-Language Models (VLMs) capable of performing OCR directly on full-scene images without explicit cropping. These models leverage joint visual-textual representations and spatial attention to identify and transcribe text in context. We consider compact VLMs that strike a balance between accuracy and resource efficiency, making them viable condidates for on-device applications. Together, these two model classes offer contrasting approaches to OCR at the edge: one relying on structured pipelines, the other on holistic scene understanding.

\subsubsection{CNN-based OCR Model}
We evaluate one widely used traditional OCR pipeline, PaddleOCRv4, focusing specifically on its recognition module. This module employs a CRNN architecture, which combines convolutional layers for feature extraction with recurrent layers for sequence modeling. This design is well-suited for recognizing variable-length text lines in natural scenes. The CRNN processes input word crops by first extracting high-level visual features with a lightweight CNN backbone, then modeling character dependencies with a bi-directional RNN, and finally decoding text sequences using CTC loss.

PaddleOCRv4 has been extensively optimized for deployment on edge and mobile devices, striking a balance between recognition accuracy and inference speed. Its modular design also allows developers to pair it with detection and layout analysis stages for end-to-end pipelines, but in this study, we isolate the recognition stage to enable a fair comparison with end-to-end Vision-Language Models. This makes PaddleOCRv4 an ideal baseline for benchmarking classic structured OCR workflows in resource-constrained settings, particularly when text regions can be reliably cropped in advance.

\subsubsection{Vision-Language Models}
We evaluate a diverse set of VLM-based architectures capable of recognizing text from full-scene images. These models integrate visual backbones with language decoders or encoders and are selected for their balance between performance and deployability on resource-constrained devices. Importantly, selected models differ in their exposure to OCR tasks during pretraining or fine-tuning - ranging from explicitly OCR-trained models to those with only partial or no known OCR-specific supervision - allowing us to assess their zero-shot and transfer capabilities.

\begin{table*}[h!]
\centering
\caption{Summary of evaluated models, including architecture type, size, and OCR training exposure. OCR Trained: \checkmark = explicitly trained, \texttildelow = partial/unclear, \xmark = no known OCR training.}
\label{tab:model-summary}
\begin{tabular}{|l|c|c|l|l|c|}
\hline
\textbf{Model} & \textbf{Type} & \textbf{Params} & \textbf{Visual Backbone} & \textbf{Language Model} & \textbf{OCR Trained} \\
\hline
PaddleOCRv4 & CNN & $\sim$15M & - & - & \checkmark \\
\hline
T-LLaVA-Gemma & VLM & 2.4B & SigLIP & Gemma-2B & \xmark \\
\hline
T-LLaVA-Phi & VLM & 3.1B & SigLIP & Phi-2 & \xmark \\
\hline
SmolVLM2 VI 256M & VLM & 256M & Custom ViT & SmolLM & \texttildelow \\
\hline
SmolVLM2 VI 500M & VLM & 500M & Custom ViT & SmolLM & \texttildelow \\
\hline
SmolVLM2 2.2B & VLM & 2.2B & Custom ViT & SmolLM & \texttildelow \\
\hline
Gemma 3 4B & VLM & 4B & ViT / SigLIP & Gemma-3 & \xmark \\
\hline
Qwen2.5 VL 3B & VLM & 3B & ViT & Qwen-2.5 & \texttildelow \\
\hline
InternVL3 1B & VLM & 1B & InternImage & InternLM & \checkmark \\
\hline
InternVL3 2B & VLM & 2B & InternImage & InternLM & \checkmark \\
\hline
LLaVA 7B & VLM & 7B & CLIP ViT-L & LLaMA-7B & \texttildelow \\
\hline
\end{tabular}
\end{table*}

The following VLMs were considered:

\begin{itemize}
    \item TinyLLaVA-Gemma-SigLIP-2.4B and TinyLLaVA-Phi-2-SigLIP-3.1B (T-LLaVA-Gemma 2.4B \& T-LLaVA-Phi 3.1B): Modular architectures that pair lightweight language models (Gemma 2B and Phi-2) with SigLIP visual encoders, enabling efficient visual understanding.
    \item SmolVLM2 Series: A family of open-weight, compact VLMs with visual insttruction-tuned capabilities. We consider the 256M, 500M and 2.2B parameter variants to explore the impact of scale on OCR tasks.
    \item Gemma 3 4B: A recent multilingual LLM integrated with a visual encoder, evaluated here for its generalization capabilities in open-scene OCR.
    \item Qwen 2.5 VL 3B: A vision-language variant of the Qwen series offering multi-modal reasoning with a compact model size.
    \item InternVL3 1B and 2B: Lightweight VLMs adapted for English OCR via instruction tuning, representing strong visual grounding with efficient parameterization. 
    \item LLaVA 7B: A larger scale VLM included for comparison, to assess the trade-offs between accuracy and inference cost on full-scene inputs. Includes some text-in-image tasks in pretraining.
\end{itemize}

A summary of all models evaluated in this study is provided in Table \ref{tab:model-summary}, including their parameter sizes, visual and language backbones, and the extent of their exposure to OCR-specific training.

\subsubsection{Evaluation Protocol}
\label{subsec:evaluation}

\paragraph{Accuracy.}
For PaddleOCRv4, we follow the standard pre-processing pipeline, including mean subtraction and normalization by standard deviation. In contrast, VLMs do not require such normalization. This is due to the fact that most modern VLMs are pretrained with visual encoders like SigLIP or CLIP, which operate on raw or normalized pixel ranges internally as part of their own vision backbone, making manual pre-processing unnecessary.

To ensure a fair comparison, two evaluation scenarios are defined:

\begin{itemize}
    \item \textbf{Cropped Text Recognition.} This setup isolates the recognition task by evaluating models on individual cropped word regions. For PaddleOCRv4, which relies on a traditional detection-recognition pipeline, only this mode is used. For VLMs, we evaluate the same cropped inputs to allow direct comparison. A prediction is considered correct only if the recognized word exactly matches the ground truth, ignoring punctuation and special characters. Minor deviations (e.g., typos or character insertions) are treated as errors.
    \item \textbf{Full Image Recognition.} In this setup, only VLMs are evaluated. Models are prompted to detect and recognize all visible words from an uncropped scene image. A prediction is counted as correct only if the model successfully identifies and transcribes all ground truth words without error; missing words or recognition inaccuracies count as incorrect. The order of recognized words is not considered.
\end{itemize}

\paragraph{Performance Proxy.}
We use the total number of parameters in each model as a proxy for computational performance. While latency and throughput vary significantly across hardware platforms and optimization toolchains, model size provides a consistent, architecture-agnostic metric for comparing deployment feasibility. In practice, larger models generally incur greater memory and compute overhead, especially on edge devices with limited resources. As most VLMs are built on standard Transformer-based backbones, performance tends to scale linearly with parameter count, making this an appropriate approximation \cite{sridhar2023instatune}\cite{aminabadi2022deepspeed}. A more detailed runtime analysis across hardware platforms is planned as part of future work.
\section{Results}

\begin{figure}
    \includegraphics[width=\columnwidth]{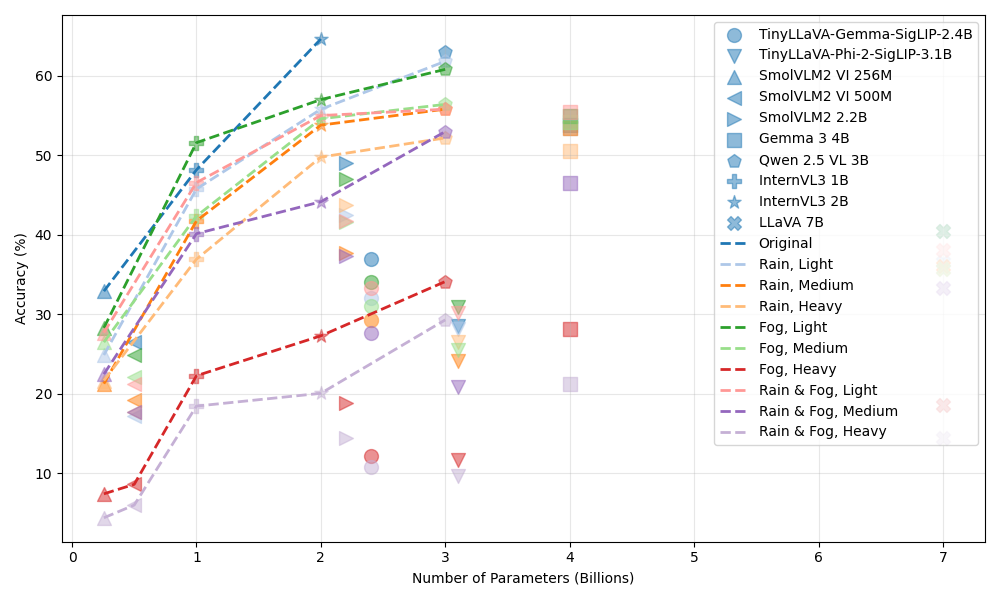}
    \caption{SVT recognition accuracy (\%) on whole images vs. model size (billions of parameters). 
Marker shapes indicate models; colors denote weather conditions and severity. 
Dashed lines connect Pareto-efficient models. Evaluation is on full-scene images without cropping.}
    \label{fig:pareto-svt}
\end{figure}

\begin{figure}
    \includegraphics[width=\columnwidth]{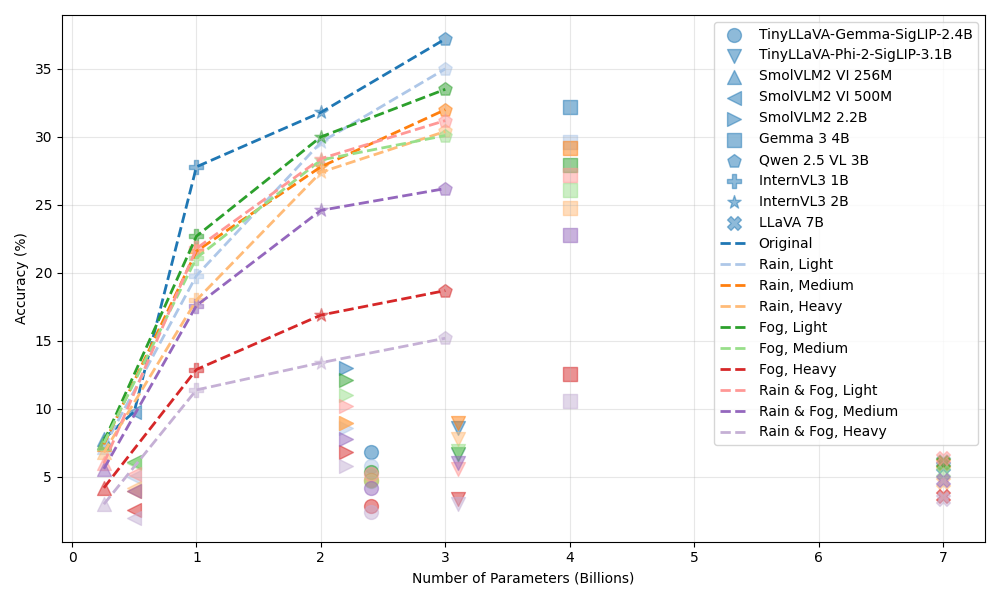}
    \caption{ICDAR 2015 recognition accuracy (\%) on whole images vs. model size (billions of parameters). 
Marker shapes indicate models; colors denote weather conditions and severity. 
Dashed lines connect Pareto-efficient models. Evaluation is on full-scene images without cropping.}
\label{fig:pareto-icdar}
\end{figure}

\begin{figure}
    \includegraphics[width=\columnwidth]{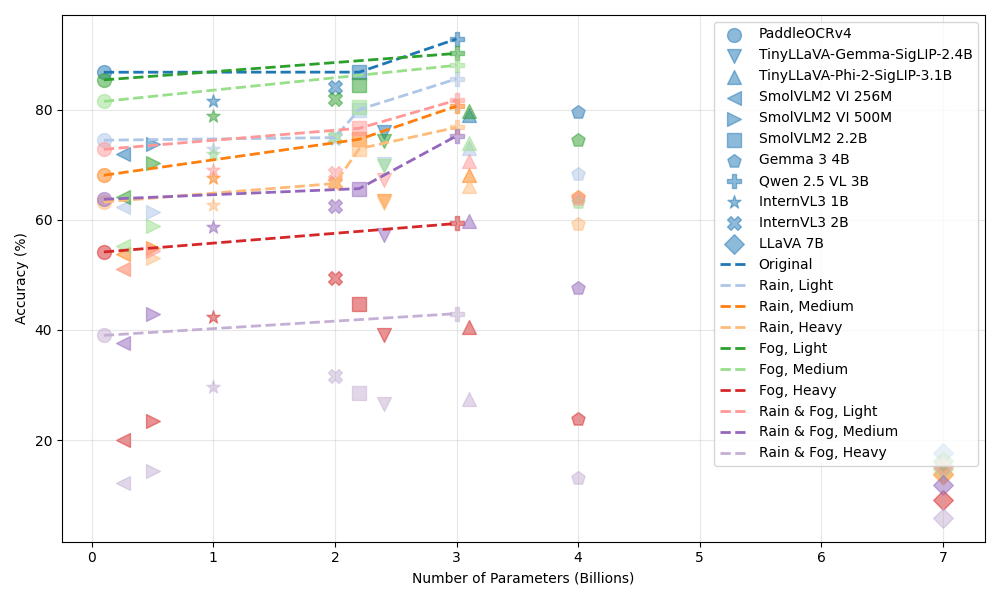}
    \caption{SVT recognition accuracy (\%) vs. model size for cropped word images. 
Marker shapes represent models; colors indicate weather conditions and severity. 
Dashed lines show Pareto-efficient performance across conditions.}
    \label{fig:pareto-svt-cropped}
\end{figure}

\begin{figure}
    \includegraphics[width=\columnwidth]{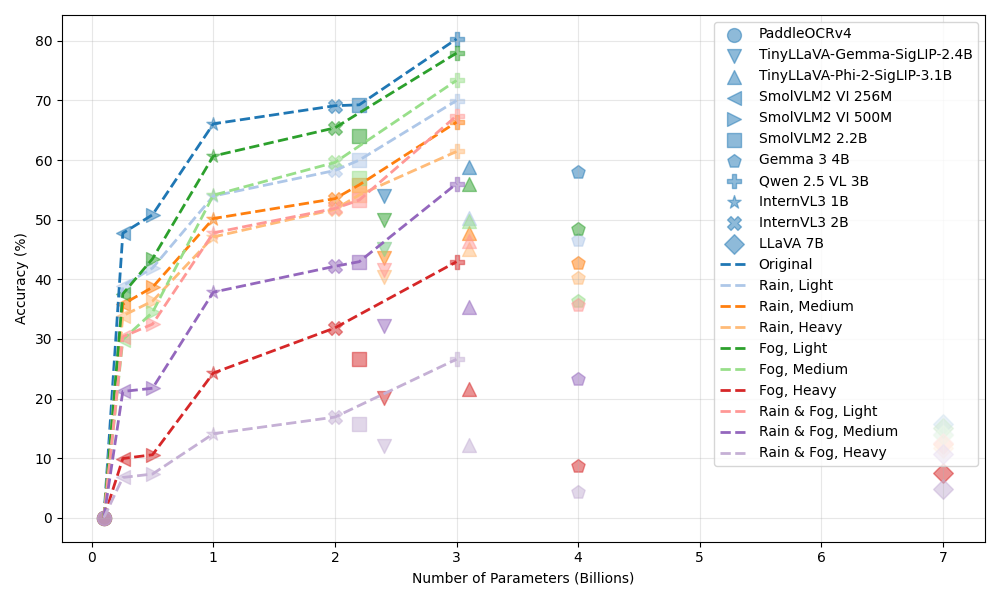}
    \caption{ICDAR 2015 recognition accuracy (\%) vs. model size for cropped word images. 
Marker shapes represent models; colors indicate weather conditions and severity. 
Dashed lines show Pareto-efficient performance across conditions.}
    \label{fig:pareto-icdar-cropped}
\end{figure}

\begin{table*}[h!]
\centering
\caption{Accuracy (\%) of VLM-based models on the SVT dataset under original and augmented weather conditions. 
Augmentations include simulated \textbf{Rain (R)}, \textbf{Fog (F)}, and \textbf{Rain+Fog (R+F)} across three severity levels: 
\textbf{Light (L)}, \textbf{Medium (M)}, and \textbf{Heavy (H)}. For example, \textbf{R-L} corresponds to light rain, 
\textbf{F-M} to medium fog, and \textbf{R+F-H} to heavy rain combined with heavy fog. 
The final column reports the average accuracy across all augmented conditions.}
\label{tab:results-svt}
\setlength{\tabcolsep}{5pt}
\begin{tabular}{|l|c|c|c|c|c|c|c|c|c|c|c|}
\hline
\textbf{Model} & \textbf{Orig} & \textbf{R-L} & \textbf{R-M} & \textbf{R} & \textbf{F-L} & \textbf{F-M} & \textbf{F-H} & \textbf{R+F-L} & \textbf{R+F-M} & \textbf{R+F-H} & \textbf{Avg.} \\
\hline
T-LLaVA-Gemma 2.4B  & 36.95 & 32.13 & 29.32 & 29.32 & 34.14 & 31.12 & 12.25 & 33.33 & 27.71 & 10.84 & 27.71\\
T-LLaVA-Phi 3.1B & 28.51 & 28.11 & 24.10 & 26.51 & 30.92 & 25.50 & 11.65 & 30.12 & 20.88 & 9.64 & 23.59 \\
SmolVLM2 VI 256M & 32.93 & 24.90 & 21.29 & 21.69 & 28.31 & 26.51 & 7.43 & 27.71 & 22.49 & 4.42 & 21.77 \\
SmolVLM2 VI 500M & 26.51 & 17.27 & 19.28 & 17.67 & 24.90 & 22.09 & 8.63 & 21.29 & 17.67 &6.02 & 18.13 \\
SmolVLM2 2.2B & 49.00 & 42.57 & 37.75 & 43.78 & 46.99 & 41.57 & 18.88 & 41.77 & 37.35 & 14.46 & 37.41 \\
Gemma 3 4B & 53.82 & 53.41 & 53.41 & 50.60 & 55.02 & 54.22 & 28.11 & 55.42 & 46.59 & 21.29 & 47.19 \\
Qwen 2.5 VL 3B & 63.05 & \textbf{61.85} & \textbf{55.83} & \textbf{52.21} & \textbf{60.84} & \textbf{56.43} & \textbf{34.14} & \textbf{55.82} & \textbf{53.01} & \textbf{29.32} & \textbf{52.25} \\
InternVL3 1B & 48.19 & 45.78 & 41.77 & 36.95 & 51.61 & 42.37 & 22.29 & 46.59 & 40.16 & 18.47 & 39.42 \\
InternVL3 2B & \textbf{64.66} & 55.82 & 53.82 & 49.80 & 57.03 & 54.62 & 27.31 & 55.02 & 44.18 & 20.08 & 48.23 \\
LLaVA 7B & 40.56 & 36.55 & 36.14 & 35.74 & 40.56 & 35.74 & 18.67 & 38.15 & 33.33 & 14.46 & 32.99 \\

\hline
\end{tabular}
\end{table*}

\begin{table*}[h]
\centering
\caption{
Recognition accuracy (\%) of VLM-based models on full-scene images from the ICDAR 2015 dataset under original and weather-augmented conditions. 
Models were prompted to detect and transcribe all visible words in the image. A prediction is counted as correct only if all ground truth words are correctly recognized. 
Abbreviations: \textbf{R = Rain}, \textbf{F = Fog}, \textbf{R+F = Rain+Fog}; \textbf{L = Light}, \textbf{M = Medium}, \textbf{H = Heavy}. The final column shows the average accuracy across all augmented conditions.
}
\label{tab:results-icdar}
\setlength{\tabcolsep}{5pt}
\begin{tabular}{|l|c|c|c|c|c|c|c|c|c|c|c|}
\hline
\textbf{Model} & \textbf{Orig} & \textbf{R-L} & \textbf{R-M} & \textbf{R} & \textbf{F-L} & \textbf{F-M} & \textbf{F-H} & \textbf{R+F-L} & \textbf{R+F-M} & \textbf{R+F-H} & \textbf{Avg.} \\
\hline
T-LLaVA-Gemma 2.4B  & 6.80 & 5.80 & 4.80 & 4.80 & 5.40 & 4.70 & 2.90 & 5.20 & 4.20 & 2.40 & 4.7\\
T-LLaVA-Phi 3.1B & 8.60 & 6.60 & 9.00 & 7.80 & 6.70 & 6.90 & 3.40 & 5.60 & 6.00 & 3.00& 4.36\\
SmolVLM2 VI 256M  & 7.80 & 7.20 & 7.40 & 6.80 & 7.60 & 7.50 & 4.20 & 6.00 & 5.60 & 3.00 & 6.31\\
SmolVLM2 VI 500M  & 9.80 & 5.00 & 4.00 & 4.20 & 6.10 & 6.00 & 2.60 & 5.20 & 4.00 & 2.00 & 4.89\\
SmolVLM2 2.2B  & 13.00 & 8.60 & 9.00 & 9.00 & 12.10 & 11.00 & 6.80 & 10.20 & 7.80 & 5.80 & 9.33\\
Gemma 3 4B & 32.20 & 29.60 & 29.20 & 24.80 & 27.90 & 26.10 & 12.60 & 27.20 & 22.80 & 10.60 & 24.3\\
Qwen 2.5 VL 3B & \textbf{37.20} & \textbf{35.00} & \textbf{32.00} & \textbf{30.40} & \textbf{33.50} & \textbf{30.10} & \textbf{18.70} & \textbf{31.20} & \textbf{26.20} & \textbf{15.20} & \textbf{28.95} \\
InternVL3 1B  & 27.80 &  19.80 & 21.60 & 18.00 & 22.70 & 21.10 & 12.90 & 21.80 & 17.60 & 11.40 & 19.47\\
InternVL3 2B  & 31.80 & 29.60 & 27.80 & 27.40 & 30.00 & 28.30 & 16.90 & 28.40 & 24.60 & 13.40    & 25.82 \\
LLaVA 7B  & 5.80 & 6.20 & 6.00 & 4.60 & 6.10 & 5.40 & 3.60 & 6.40 & 4.80 & 3.40 & 5.22 \\
\hline
\end{tabular}
\end{table*}

\begin{table*}[h]
\centering
\caption{
Recognition accuracy (\%) of all models, including PaddleOCRv4, on cropped word regions from the SVT dataset under original and weather-augmented conditions. 
Each cropped image contains a single word, and models are evaluated on exact transcription accuracy (ignoring punctuation and special characters). 
Abbreviations: \textbf{R = Rain}, \textbf{F = Fog}, \textbf{R+F = Rain+Fog}; \textbf{L = Light}, \textbf{M = Medium}, \textbf{H = Heavy}. The final column shows the average accuracy across all augmented conditions.
}
\label{tab:results-svt-cropped}
\setlength{\tabcolsep}{5pt}
\begin{tabular}{|l|c|c|c|c|c|c|c|c|c|c|c|}
\hline
\textbf{Model} & \textbf{Orig} & \textbf{R-L} & \textbf{R-M} & \textbf{R} & \textbf{F-L} & \textbf{F-M} & \textbf{F-H} & \textbf{R+F-L} & \textbf{R+F-M} & \textbf{R+F-H} & \textbf{Avg.} \\
\hline
PaddleOCRv4 & 86.84 & 74.49 & 68.12 & 63.27 & 85.45 & 81.55 & 54.18 & 72.83 & 63.75 & 39.02 & 68.95\\
\hline
T-LLaVA-Gemma 2.4B  & 76.04 & 70.17 & 63.52 & 63.06 & 74.34 & 69.86 & 39.10 & 67.23 & 57.19 & 26.58 &60.71 \\
T-LLaVA-Phi 3.1B & 79.13 & 73.11 & 68.16 & 66.15 & 79.75 & 74.03 & 40.49 & 70.63 & 59.81 & 27.51 & 63.88\\
SmolVLM2 VI 256M & 72.02 & 62.29 & 53.79 & 51.16 & 64.14 & 55.33 & 19.94 & 51.00 & 37.71 & 12.21 & 47.96 \\
SmolVLM2 VI 500M & 73.72 & 61.36 & 54.87 & 53.01 & 70.32 & 58.89 & 23.49 & 54.40 & 42.97 & 14.37 & 50.74 \\
SmolVLM2 2.2B & 86.86 & 80.06 & 74.65 & 72.95 & 84.54 & 80.53 & 44.67 & 76.66 & 65.69 & 28.59 & 69.52 \\
Gemma 3 4B & 79.60 & 68.32 & 64.14 & 59.20 & 74.50 & 63.21 & 23.80 & 63.83 & 47.60 & 13.14 & 55.73\\
Qwen 2.5 VL 3B & \textbf{92.89} & \textbf{85.63} & \textbf{80.68} & \textbf{76.82} & \textbf{90.26} & \textbf{88.10} & \textbf{59.35} & \textbf{81.76} & \textbf{75.27} & \textbf{42.97} & \textbf{77.373}\\
InternVL3 1B & 81.61 & 72.95 & 67.54 & 62.75 & 78.83 & 72.02 & 42.35 & 69.09 & 58.73 & 29.68 & 63.55\\
InternVL3 2B & 84.23 & 74.96 & 66.77 & 66.62 & 81.92 & 75.12 & 49.46 & 68.47 & 62.44 & 31.68 & 66.17\\
LLaVA 7B & 17.62 & 15.46 & 13.91 & 13.60 & 16.23 & 14.68 & 9.12 & 15.15 & 11.75 & 5.87 & 13.34\\

\hline
\end{tabular}
\end{table*}

\begin{table*}[h]
\centering
\caption{
Recognition accuracy (\%) of all models, including PaddleOCRv4, on cropped word regions from the ICDAR 2015 dataset under original and weather-augmented conditions. 
Each cropped image contains a single word, and models are evaluated based on exact string matching with the ground truth (ignoring punctuation and special characters). 
Abbreviations: \textbf{R = Rain}, \textbf{F = Fog}, \textbf{R+F = Rain+Fog}; \textbf{L = Light}, \textbf{M = Medium}, \textbf{H = Heavy}. The final column shows the average accuracy across all augmented conditions.
}
\label{tab:results-icdar-cropped}
\setlength{\tabcolsep}{5pt}
\begin{tabular}{|l|c|c|c|c|c|c|c|c|c|c|c|}
\hline
\textbf{Model} & \textbf{Orig} & \textbf{R-L} & \textbf{R-M} & \textbf{R} & \textbf{F-L} & \textbf{F-M} & \textbf{F-H} & \textbf{R+F-L} & \textbf{R+F-M} & \textbf{R+F-H} & \textbf{Avg.} \\
\hline
PaddleOCRv4 & 76.83 & 66.32 & 61.89 & 56.48 & 74.78 & 71.44 & \textbf{46.58} & 63.76 & 55.45 & \textbf{33.26} & 60.68 \\
\hline
T-LLaVA-Gemma 2.4B  & 54.01 & 45.03 & 43.58 & 40.30 & 49.86 & 45.03 & 20.03 & 41.55 & 32.24 & 11.97 & 38.36 \\
T-LLaVA-Phi 3.1B & 58.83 & 50.34 & 47.73 & 45.13 & 55.94 & 49.71 & 21.57 & 46.38 & 35.42 & 12.16& 42.32 \\
SmolVLM2 VI 256M  & 47.73 & 38.90 & 35.81 & 33.88 & 37.55 & 29.87 & 9.94 & 30.36 & 21.19 & 6.76 & 29.2 \\
SmolVLM2 VI 500M  & 50.82 & 41.89 & 38.66 & 36.34 & 43.44 & 34.46 & 10.57 & 32.48 & 21.72 & 7.34 & 31.77 \\
SmolVLM2 2.2B  & 69.26 & 59.99 & 55.84 & 54.15 & 64.09 & 57.00 & 26.69 & 53.28 & 42.95 & 15.69 & 49.89 \\
Gemma 3 4B  & 58.06 & 46.57 & 42.81 & 40.15 & 48.41 & 36.44 & 8.69 & 35.71 & 23.31 & 4.39 & 34.45 \\
Qwen 2.5 VL 3B  & \textbf{80.31} & \textbf{69.98} & \textbf{66.36} & \textbf{61.44} & \textbf{77.94} & \textbf{73.36} & 42.95 & \textbf{67.42} & \textbf{56.03} & 26.59 & \textbf{62.24} \\
InternVL3 1B  & 66.07 & 53.96 & 50.19 & 47.10 & 60.67 & 54.10 & 24.28 & 47.78 & 37.84 & 14.09 & 45.61 \\
InternVL3 2B  & 69.11 & 58.25 & 53.52 & 51.74 & 65.40 & 59.60 & 31.85 & 51.88 & 42.18 & 16.89 & 50.04 \\
LLaVA 7B  & 15.69 & 14.00 & 12.45 & 11.73 & 15.06 & 13.85 & 7.58 & 12.60 & 10.76 & 4.87 & 11.86 \\
\hline
\end{tabular}
\end{table*}


This section presents the performance of selected OCR models across original and augmented datasets, evaluating both full-scene and cropped image inputs. We report recognition accuracy (\%) and analyze the impact of weather-induced distortions: rain, fog, and their combination, on model robustness. Model size (in billions of parameters) is used as a proxy for computational footprint and edge deployment feasibility.

\subsection{Full-Scene Recognition}

VLMs were tested on full, unprocessed images from the SVT and ICDAR 2015datasets. As shown in Tables \ref{tab:results-svt} and \ref{tab:results-icdar}, models experienced varying degrees of degradation under weather augmentations, with heavy fog and rain+fog proving the most challenging.

Across both datasets, \textbf{Qwen 2.5 VL 3B} consistently demonstrated the highest overall robustness, achieving top performance in original and distorted conditions. \textbf{InternVL3 2B} and \textbf{Gemma 3 4B} also performed strongly but showed greater sensitivity to visual noise. Smaller models, such as \textbf{SmolVLM2 VI 256M}, exhibited sharper drops in accuracy, underscoring a performance-resource trade-off. Pareto curves in Figures \ref{fig:pareto-svt} and \ref{fig:pareto-icdar} highlight the most efficient models under each condition.

\subsubsection{Cropped Recognition}

Cropped evaluation results are reported in Tables \ref{tab:results-svt-cropped} and \ref{tab:results-icdar-cropped}. This setup isolates the recognition task by removing the need for detection, enabling a direct comparison between VLMs and the traditional CNN-based OCR approach.

PaddleOCRv4, designed explicitly for cropped scene text recognition, performed strongly, achieving competitive accuracy across both datasets while maintaining a much smaller model size. On the cropped ICDAR dataset, it even outperformed all VLMs under the heaviest weather conditions, highlighting the robustness of its architecture and training.

Among the VLMs, Qwen 2.5 VL 3B consistently delivered the highest accuracy overall, trailing PaddleOCRv4 only in a few challenging cases. SmolVLM2 2.2B also demonstrated a strong balance of size and accuracy, particularly on ICDAR, suggesting its practicality for edge deployments with moderate resource constraints.


\section{Conclusion}
This study compared traditional CNN-based OCR with modern VLMs for billboard text recognition in realistic outdoor conditions. Results show that OCR accuracy consistently degrades under increasing weather severity, highlighting the need for robust models in real-world deployment.

Among the VLMs, models with explicit or partial OCR supervision - such as InternVL3 and Qwen 2.5 VL 3B - were the most resilient to visual noise, with Qwen 2.5 VL 3B emerging as the strongest overall performer for full-scene recognition. However, for cropped text regions, PaddleOCRv4 remained highly competitive, matching or surpassing larger VLMs while requiring only a fraction of the parameters.

These results illustrate a practical trade-off: VLMs offer valuable whole-image context and flexible scene reasoning but at the cost of higher latency and energy use. For resource-constrained edge devices, structured pipelines like PaddleOCRv4 still provide excellent recognition accuracy and efficiency when text regions can be reliably detected.

To support further research on OCR robustness, the weather-augmented versions of ICDAR and SVT used in this study are released publicly. Future work will extend this benchmark with integrated detection pipelines, deeper ViT-based exploration, and real-world edge deployment tests.

\section{References}

{
    \small
    \bibliographystyle{ieeenat_fullname}
    \bibliography{main}
}


\end{document}